\documentclass[11pt]{article}

\usepackage[preprint]{acl}

\usepackage{times}
\usepackage{latexsym}
\usepackage{tikz}
\usepackage{pgfplots}
\pgfplotsset{compat=1.18} 
\usepgfplotslibrary{groupplots}
\usepackage[most]{tcolorbox}
\tcbuselibrary{breakable,skins}

\usepackage{listings}
\usepackage{enumitem}

\usepackage{xcolor}

\usepackage{pgfplots}
\usepgfplotslibrary{groupplots}
\pgfplotsset{compat=1.18}
\usepackage{placeins} 
\usepackage{capt-of}

\usepackage[T1]{fontenc}

\usepackage[utf8]{inputenc}

\usepackage{microtype}

\usepackage{inconsolata}

\usepackage{graphicx}
\usepackage{amsmath}
\usepackage{multirow}
\usepackage{soul}
\usepackage{booktabs}
\usepackage{multirow}
\usepackage{makecell}

\usepackage[table]{xcolor}  
\definecolor{headergray}{gray}{0.92}

\usepackage{booktabs}   
\usepackage{tabularx}   
\usepackage{makecell}   
\usepackage{array}
\newcolumntype{Y}{>{\raggedright\arraybackslash}X}
\usepackage{multirow}   
\usepackage{adjustbox}
\usepackage{todonotes}

\title{Experiments or Outcomes? Probing Scientific Feasibility in Large Language Models}

\author{Seyedali Mohammadi, Manas Gaur, Francis Ferraro \\
        University of Maryland, Baltimore County, MD, USA \\ 
        \{m294,manas, ferraro\}@umbc.edu}

\begin{document}
\maketitle

\begin{abstract}
Scientific feasibility assessment asks whether a claim is consistent with established knowledge and whether experimental evidence could support or refute it. We frame feasibility assessment as a diagnostic reasoning task in which, given a hypothesis, a model predicts $\textsc{feasible}$ or $\textsc{infeasible}$ and justifies its decision. We evaluate large language models (LLMs) under controlled knowledge conditions (hypothesis-only, with experiments, with outcomes, or both) and probe robustness by progressively removing portions of the experimental and/or outcome context. Across multiple LLMs and two datasets, providing outcome evidence is generally more reliable than providing experiment descriptions. Outcomes tend to improve accuracy beyond what internal knowledge alone provides, whereas experimental text can be brittle and may degrade performance when the context is incomplete. These findings clarify when experimental evidence benefits LLM-based feasibility assessment and when it introduces fragility.
\end{abstract}

\section{Introduction}
\label{introduction}
While LLMs can be deployed in scientific workflows, from literature review to hypothesis generation and experimental planning~\citep{eger2025transformingsciencelargelanguage,zheng-etal-2025-automation}, their capacity to perform a fundamental scientific task remains poorly understood. \textit{\textbf{Scientific feasibility assessment}} determines whether a claim aligns with established knowledge and admits concrete experiments that could support or refute it. Feasibility assessment requires diagnostic reasoning about experimental designs, expected outcomes, and evidential alignment.

\begin{figure}[!t]
    \raggedleft
    \includegraphics[width=0.99\linewidth]{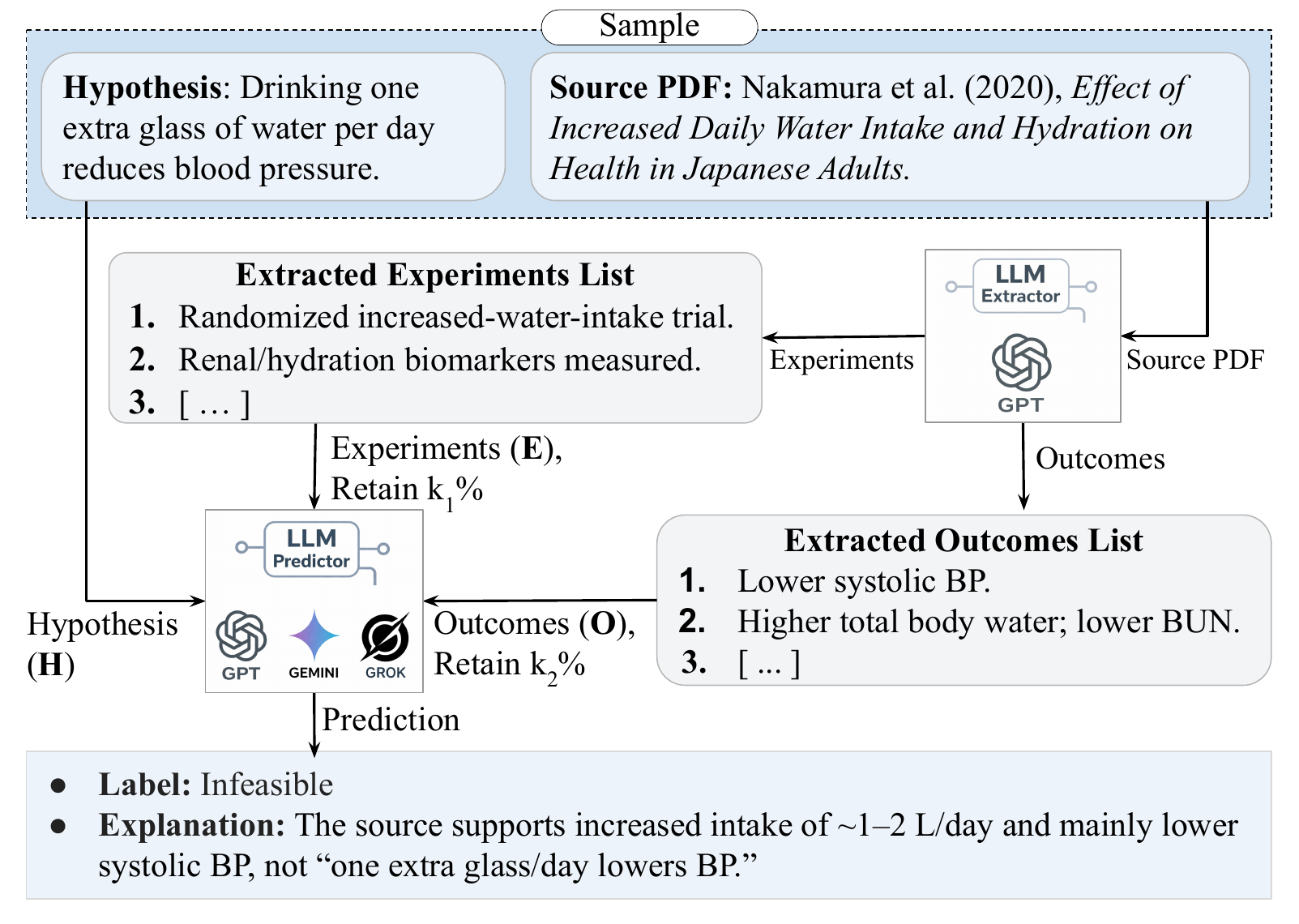}
    \caption{Controlled evidence framework for scientific feasibility assessment. Given a hypothesis, extracted experiments and outcomes from a source paper are revealed independently, where $k_1$ controls the fraction of experiments shown and $k_2$ controls the fraction of outcomes shown. The model predicts feasibility based on the provided evidence. For illustration, the example hypothesis and source paper shown here are both drawn from \citet{nu12041191}.\protect\footnotemark}
\label{fig:motivation}
\end{figure}
\footnotetext{LLM Predictor and LLM Extractor icons were created with ChatGPT assistance.}

Figure~\ref{fig:motivation} illustrates a claim (\textit{``Drinking one extra glass of water per day reduces blood pressure.''}) where feasibility hinges on evaluating specific tests (e.g., water-supplementation randomized control trials (RCTs)) and whether their outcomes match the hypothesized effect. Even with relevant trials, uncertainty can persist due to mixed results, dose mismatches (one glass vs.\ liters), and varying endpoints. This shows that feasibility depends on \textit{which} experiments and outcomes are considered, not merely on retrieving related papers.

Despite growing interest in LLMs for scientific tasks, existing work either focuses on hypothesis generation rather than assessment~\cite{qi2023large, yang-etal-2024-large-language,LIU2025121307}, combines internal model knowledge with retrieval without isolating when each is sufficient~\cite{jansen-etal-2025-matter,rao-etal-2025-nsf}, or examines adherence to externally provided knowledge (label definitions) in non-scientific settings~\citep{mohammadi-etal-2025-llms}. These leave three critical questions unanswered: (\textbf{RQ1}) Can LLMs assess feasibility using only parametric knowledge? (\textbf{RQ2}) How does providing explicit experimental context, experiments alone, outcomes alone, or both together, change feasibility judgments compared to internal knowledge? (\textbf{RQ3}) How robust are these judgments when experimental information is incomplete? We address these questions through a controlled knowledge framework that systematically varies what context accompanies a hypothesis while holding the prediction task constant. 
%
%

We make two primary contributions: (1) a controlled-knowledge framework to measure how experiments and outcomes shift feasibility judgments, and (2) a stability analysis quantifying robustness under omissions. We find that aligned evidence can improve accuracy, but partial evidence often hurts, sometimes dropping below the hypothesis-only baseline, and degradation is frequently non-monotonic, indicating brittle, superficially aligned reasoning rather than graceful uncertainty handling. To support reproducibility, our code, prompts, and evaluation scripts are in this repository: \url{https://github.com/mohammadi-ali/scify}.


\section{Problem Formulation}
\label{problem_formulation}
We formulate \emph{scientific feasibility assessment} as a structured prediction problem under \emph{controlled knowledge} about a hypothesis.
Let $h$ denote a scientific hypothesis (claim). Each instance is annotated with a ground-truth feasibility label
$y^\star \in \mathcal{Y}$, where $\mathcal{Y}=\{\textsc{feasible}, \textsc{infeasible}\}$. 
When available, the instance also includes a set of \emph{source experiments} $\mathcal{E}^\star$ and their \emph{reported outcomes}
$\mathcal{O}^\star$, extracted from the source paper provided as part of the dataset (i.e., not retrieved). 
We use a binary formulation as a controlled diagnostic setting rather than a complete model of scientific judgment. This choice also matches the annotation schema of the dataset, which provides claim-level $\textsc{feasible}/\textsc{infeasible}$ labels.

\paragraph{Model prediction}
Given a hypothesis $h$ and an optional context $x$, an LLM $f_\theta$ with parameters $\theta$ outputs both a label $\hat{y} \in \mathcal{Y}$ and a brief justification $\hat{e}$. The context $x$ is constructed to control which parts of experimental evidence are provided.

\textbf{Controlled Knowledge Framework:} We evaluate models under four conditions that vary the optional context $x$ provided alongside hypothesis $h$: ~\textbf{H (Hypothesis-only):} $x = \emptyset$. The model relies exclusively on parametric knowledge. This isolates the baseline capability without external evidence and serves as the reference point for all augmented settings.~\textbf{H+E (Hypothesis + Experiments):} $x = \mathcal{E}^*$. The model receives experiment descriptions but \emph{not} their outcomes. This tests whether models can reason about study designs and infer potential outcomes, or whether they require explicit results. ~\textbf{H+O (Hypothesis + Outcomes):} $x = \mathcal{O}^*$. The model receives outcome summaries but \emph{not} the experimental procedures that generated them. This tests whether outcome statements alone provide sufficient evidential grounding, or whether experimental context is necessary for interpretation.~\textbf{H+E+O (Hypothesis + Experiments + Outcomes):} $x = (\mathcal{E}^*, \mathcal{O}^*)$. The model receives a complete experimental context. This represents the ideal condition where study designs and results are both available. In all four settings, the prediction task is \emph{identical}: output $(\hat{y}, \hat{e})$ for hypothesis $h$. Only the provided context $x$ varies. This controlled design enables direct comparison: any difference in predictions or justifications across settings reflects the impact of experimental evidence, not task variation. By holding the task constant and varying only $x \in \{\emptyset, \mathcal{E}^*, \mathcal{O}^*, (\mathcal{E}^*, \mathcal{O}^*)\}$, we can answer our \textbf{RQs}. 

\textbf{Stability Analysis to Test Robustness Under Partial Information}: Real scientific reasoning often proceeds with incomplete evidence. We introduce a \emph{stability analysis} that progressively removes portions of provided experiments and outcomes, measuring how gracefully model judgments degrade. This reveals whether models exhibit (i)~monotonic degradation (performance decreases smoothly as evidence decreases), or (ii)~brittle collapse (performance drops sharply or non-monotonically), indicating over-reliance on specific evidence pieces.

\textbf{Partial Revelation Parameters.} Let $k_1 \in [0, 1]$ denote the fraction of experiments revealed and $k_2 \in [0, 1]$ the fraction of outcomes revealed. We evaluate at three levels, $k_1, k_2 \in \{0,  0.5, 1.0\}$. For a given $(k_1, k_2)$ pair, the provided context becomes
$x_{k_1, k_2} = (\mathcal{E}_{k_1}, \mathcal{O}_{k_2})$
where $\mathcal{E}_{k_1} \subseteq \mathcal{E}^*$ with $|\mathcal{E}_{k_1}| = \lfloor k_1 \cdot |\mathcal{E}^*| \rfloor$ and similarly for $\mathcal{O}_{k_2}$.

\paragraph{Sampling Strategy.} When $k_1 < 1$ or $k_2 < 1$, we sample subsets uniformly at random without replacement. To account for sampling variance:
(a) For each instance $h$ and each $(k_1, k_2)$ configuration, we generate $R = 5$ independent random samples. (b) We report mean performance and standard deviation across the $R$ samples. (c) Random seeds are fixed for reproducibility (see Appendix~\ref{app:implementation}).


\paragraph{Special Cases.} The $(k_1,k_2)$ framework subsumes our four settings: H has $(k_1,k_2)=(0,0)$ with $x=\emptyset$; H+E has $(1,0)$ with $x=\mathcal{E}^*$; H+O has $(0,1)$ with $x=\mathcal{O}^*$; and H+E+O has $(1,1)$ with $x=(\mathcal{E}^*,\mathcal{O}^*)$.

\paragraph{Stability Metrics.} For each dataset, we compute: (a) \textbf{Degradation curve}: Accuracy/MCC as a function of revealing level; (b) \textbf{Below-baseline rate}: Fraction of $(k_1, k_2)$ configurations where performance falls below $k_1 = k_2 = 0$ (H baseline). Non-monotonic degradation (e.g., $k_1 = 0.5$ performing worse than $k_1 = 0$) or performance below H baseline indicates that partial evidence actively \emph{misleads} rather than helps, suggesting superficial alignment rather than robust reasoning.

\paragraph{Evaluation.}
We evaluate models along two main dimensions: feasibility labels and natural-language explanations. For examples with ground-truth feasibility labels, we report overall accuracy, macro-averaged F$_1$ across the two classes ($\textsc{feasible}$ or $\textsc{infeasible}$), and Matthews correlation coefficient (MCC), which is more informative under class imbalance. For the \textsc{Matter-of-Fact} dataset that also includes gold explanations, we use a lightweight comparison between the model’s justification and the reference explanation based on lexical overlap. We emphasize that this measure is not intended to assess the logical validity or scientific soundness of the reasoning. Rather, it serves only as a diagnostic signal of how explanation content shifts across evidence conditions.

\section{Setup and Methodology}
\label{methodology}

\noindent \textbf{Models:} We evaluate several proprietary LLMs across
different capability tiers and vendors to test whether our findings are robust
to model scale and design. Specifically, we use \texttt{gpt-5.1} and
\texttt{gpt-4o}, Gemini-2.5-Pro (\texttt{gem-pro}) and Gemini-2.5-Flash
(\texttt{gem-flash}), and Grok-4.1-fast (\texttt{grok}). This selection enables
two controlled comparisons: (i) frontier versus efficiency-tier models within
the same vendor, and (ii) consistency of feasibility judgments across vendors.
All models are prompted to produce a feasibility label and a brief justification
using identical task instructions.

\noindent \textbf{Datasets:} Our study focuses on benchmarks where feasibility judgments depend
on the relationship between a hypothesis and structured scientific evidence,
rather than on surface-level factual correctness or citation matching. We
therefore select datasets whose evidence structure supports feasibility
reasoning under different levels of uncertainty.


\newcommand{\numcell}[1]{\makebox[5em][l]{#1}} 
\newcommand{\dashcell}{\numcell{--}}
\newcommand{\stdcell}[2]{\numcell{#1$\pm$#2}}

\begin{table*}[!ht]
\centering
\adjustbox{max width=0.90\textwidth}{
\small
\begin{tabular}{lll cccc cc}
\toprule
\rowcolor{gray!20}
\textbf{Model} & \textbf{R(\%)} & \textbf{Scenario} &
\multicolumn{4}{c}{\textbf{\textsc{MoF}}} &
\multicolumn{2}{c}{\textbf{\textsc{Reasons}}} \\
\cmidrule(lr){4-7}\cmidrule(lr){8-9}
\rowcolor{gray!20}
 &  &  & $Acc$ & $F1_{\text{macro}}$ & $MCC$ & $Rouge$ & $Acc$ & $F1_{\text{macro}}$ \\
\midrule

\multirow{7}{*}{\texttt{gpt-5.1}}
& 0   & \textbf{H}       & \stdcell{0.68}{0.03} & \stdcell{0.67}{0.03} & \stdcell{0.42}{0.05} & \stdcell{0.21}{0.01} & \stdcell{0.84}{0.01} & \stdcell{0.46}{0.00} \\
\cmidrule(lr){2-9}
& \multirow{3}{*}{$50$}
& \textbf{H+E}   & \stdcell{\textbf{0.67}}{\textbf{0.04}} & \stdcell{0.65}{0.04} & \stdcell{\textbf{0.41}}{\textbf{0.06}} & \stdcell{0.19}{0.00} & \stdcell{0.85}{0.04} & \stdcell{0.44}{0.01} \\
&  & \textbf{H+O}   & \stdcell{\textbf{0.67}}{\textbf{0.05}} & \stdcell{\textbf{0.67}}{\textbf{0.05}} & \stdcell{0.35}{0.10} & \stdcell{\textbf{0.21}}{\textbf{0.00}} & \stdcell{\textbf{0.91}}{\textbf{0.02}} & \stdcell{\textbf{0.47}}{\textbf{0.01}} \\
&  & \textbf{H+E+O} & \stdcell{0.65}{0.05} & \stdcell{\textbf{0.67}}{\textbf{0.05}} & \stdcell{0.32}{0.09} & \stdcell{0.20}{0.00} & \stdcell{0.90}{0.02} & \stdcell{\textbf{0.47}}{\textbf{0.01}} \\
\cmidrule(lr){2-9}
& \multirow{3}{*}{$100$}
& \textbf{H+E}   & \stdcell{\textbf{0.70}}{\textbf{0.05}} & \stdcell{\textbf{0.69}}{\textbf{0.05}} & \stdcell{\textbf{0.44}}{\textbf{0.08}} & \stdcell{0.19}{0.00} & \stdcell{0.84}{0.02} & \stdcell{0.45}{0.01} \\
&  & \textbf{H+O}   & \stdcell{0.66}{0.04} & \stdcell{0.66}{0.04} & \stdcell{0.33}{0.07} & \stdcell{\textbf{0.21}}{\textbf{0.01}} & \stdcell{0.92}{0.01} & \stdcell{\textbf{0.47}}{\textbf{0.00}} \\
&  & \textbf{H+E+O} & \stdcell{0.66}{0.03} & \stdcell{0.66}{0.03} & \stdcell{0.33}{0.06} & \stdcell{0.20}{0.01} & \stdcell{\textbf{0.93}}{\textbf{0.02}} & \stdcell{\textbf{0.47}}{\textbf{0.01}} \\
\midrule

\multirow{7}{*}{\texttt{gpt-4o}}
& 0   & \textbf{H}       & \stdcell{0.60}{0.04} & \stdcell{0.55}{0.04} & \stdcell{0.32}{0.02} & \stdcell{0.22}{0.00} & \stdcell{0.79}{0.02} & \stdcell{0.44}{0.01} \\
\cmidrule(lr){2-9}
& \multirow{3}{*}{$50$}
& \textbf{H+E}   & \stdcell{0.52}{0.04} & \stdcell{0.44}{0.04} & \stdcell{0.25}{0.04} & \stdcell{0.18}{0.01} & \stdcell{\textbf{0.99}}{\textbf{0.01}} & \stdcell{\textbf{0.49}}{\textbf{0.00}} \\
&  & \textbf{H+O}   & \stdcell{\textbf{0.61}}{\textbf{0.03}} & \stdcell{\textbf{0.64}}{\textbf{0.03}} & \stdcell{\textbf{0.26}}{\textbf{0.05}} & \stdcell{\textbf{0.22}}{\textbf{0.01}} & \stdcell{0.93}{0.01} & \stdcell{0.48}{0.00} \\
&  & \textbf{H+E+O} & \stdcell{\textbf{0.61}}{\textbf{0.04}} & \stdcell{0.63}{0.03} & \stdcell{0.25}{0.07} & \stdcell{\textbf{0.22}}{\textbf{0.01}} & \stdcell{0.97}{0.01} & \stdcell{\textbf{0.49}}{\textbf{0.00}} \\
\cmidrule(lr){2-9}
& \multirow{3}{*}{$100$}
& \textbf{H+E}   & \stdcell{0.55}{0.04} & \stdcell{0.47}{0.03} & \stdcell{0.26}{0.03} & \stdcell{0.17}{0.00} & \stdcell{\textbf{0.98}}{\textbf{0.01}} & \stdcell{\textbf{0.50}}{\textbf{0.00}} \\
&  & \textbf{H+O}   & \stdcell{\textbf{0.64}}{\textbf{0.04}} & \stdcell{\textbf{0.64}}{\textbf{0.04}} & \stdcell{\textbf{0.30}}{\textbf{0.07}} & \stdcell{\textbf{0.22}}{\textbf{0.00}} & \stdcell{0.93}{0.01} & \stdcell{0.48}{0.00} \\
&  & \textbf{H+E+O} & \stdcell{\textbf{0.64}}{\textbf{0.03}} & \stdcell{\textbf{0.64}}{\textbf{0.03}} & \stdcell{\textbf{0.30}}{\textbf{0.06}} & \stdcell{0.21}{0.00} & \stdcell{0.96}{0.01} & \stdcell{0.49}{0.00} \\
\midrule

\multirow{7}{*}{\texttt{gem-pro}}
& 0   & \textbf{H}       & \stdcell{0.67}{0.03} & \stdcell{0.65}{0.03} & \stdcell{0.42}{0.03} & \stdcell{0.22}{0.01} & \stdcell{0.87}{0.04} & \stdcell{0.47}{0.01} \\
\cmidrule(lr){2-9}
& \multirow{3}{*}{$50$}
& \textbf{H+E}   & \stdcell{0.48}{0.04} & \stdcell{0.43}{0.03} & \stdcell{0.11}{0.04} & \stdcell{0.20}{0.01} & \stdcell{0.90}{0.05} & \stdcell{\textbf{0.47}}{\textbf{0.01}} \\
&  & \textbf{H+O}   & \stdcell{\textbf{0.69}}{\textbf{0.05}} & \stdcell{0.67}{0.05} & \stdcell{\textbf{0.40}}{\textbf{0.09}} & \stdcell{\textbf{0.22}}{\textbf{0.01}} & \stdcell{0.89}{0.07} & \stdcell{0.46}{0.02} \\
&  & \textbf{H+E+O} & \stdcell{0.66}{0.06} & \stdcell{\textbf{0.68}}{\textbf{0.06}} & \stdcell{0.34}{0.12} & \stdcell{\textbf{0.22}}{\textbf{0.01}} & \stdcell{\textbf{0.91}}{\textbf{0.07}} & \stdcell{0.46}{0.02} \\
\cmidrule(lr){2-9}
& \multirow{3}{*}{$100$}
& \textbf{H+E}   & \stdcell{0.53}{0.04} & \stdcell{0.43}{0.04} & \stdcell{0.17}{0.04} & \stdcell{0.19}{0.01} & \stdcell{\textbf{0.91}}{\textbf{0.05}} & \stdcell{\textbf{0.47}}{\textbf{0.01}} \\
&  & \textbf{H+O}   & \stdcell{\textbf{0.66}}{\textbf{0.04}} & \stdcell{\textbf{0.66}}{\textbf{0.04}} & \stdcell{\textbf{0.33}}{\textbf{0.08}} & \stdcell{\textbf{0.22}}{\textbf{0.01}} & \stdcell{\textbf{0.91}}{\textbf{0.13}} & \stdcell{0.45}{0.04} \\
&  & \textbf{H+E+O} & \stdcell{\textbf{0.66}}{\textbf{0.04}} & \stdcell{\textbf{0.66}}{\textbf{0.04}} & \stdcell{\textbf{0.33}}{\textbf{0.07}} & \stdcell{\textbf{0.22}}{\textbf{0.00}} & \stdcell{\textbf{0.91}}{\textbf{0.11}} & \stdcell{0.45}{0.03} \\
\midrule

\multirow{7}{*}{\texttt{gem-flash}}
& 0   & \textbf{H}       & \stdcell{0.64}{0.04} & \stdcell{0.60}{0.03} & \stdcell{0.39}{0.05} & \stdcell{0.21}{0.00} & \stdcell{0.77}{0.02} & \stdcell{0.42}{0.01} \\
\cmidrule(lr){2-9}
& \multirow{3}{*}{$50$}
& \textbf{H+E}   & \stdcell{0.53}{0.03} & \stdcell{0.51}{0.03} & \stdcell{0.23}{0.02} & \stdcell{0.19}{0.01} & \stdcell{\textbf{0.90}}{\textbf{0.04}} & \stdcell{\textbf{0.46}}{\textbf{0.01}} \\
&  & \textbf{H+O}   & \stdcell{\textbf{0.67}}{\textbf{0.04}} & \stdcell{\textbf{0.67}}{\textbf{0.04}} & \stdcell{\textbf{0.35}}{\textbf{0.07}} & \stdcell{\textbf{0.22}}{\textbf{0.01}} & \stdcell{0.86}{0.03} & \stdcell{0.45}{0.01} \\
&  & \textbf{H+E+O} & \stdcell{0.66}{0.05} & \stdcell{\textbf{0.67}}{\textbf{0.05}} & \stdcell{\textbf{0.35}}{\textbf{0.09}} & \stdcell{0.21}{0.01} & \stdcell{0.89}{0.03} & \stdcell{\textbf{0.46}}{\textbf{0.01}} \\
\cmidrule(lr){2-9}
& \multirow{3}{*}{$100$}
& \textbf{H+E}   & \stdcell{0.57}{0.04} & \stdcell{0.50}{0.03} & \stdcell{0.27}{0.03} & \stdcell{0.18}{0.01} & \stdcell{\textbf{0.91}}{\textbf{0.02}} & \stdcell{\textbf{0.47}}{\textbf{0.00}} \\
&  & \textbf{H+O}   & \stdcell{\textbf{0.66}}{\textbf{0.04}} & \stdcell{\textbf{0.66}}{\textbf{0.04}} & \stdcell{\textbf{0.33}}{\textbf{0.07}} & \stdcell{\textbf{0.22}}{\textbf{0.01}} & \stdcell{0.87}{0.02} & \stdcell{0.46}{0.00} \\
&  & \textbf{H+E+O} & \stdcell{0.65}{0.04} & \stdcell{0.65}{0.04} & \stdcell{0.32}{0.07} & \stdcell{0.21}{0.01} & \stdcell{0.89}{0.01} & \stdcell{\textbf{0.47}}{\textbf{0.00}} \\
\midrule

\multirow{7}{*}{\texttt{grok}}
& 0   & \textbf{H}       & \stdcell{0.65}{0.04} & \stdcell{0.63}{0.05} & \stdcell{0.36}{0.06} & \stdcell{0.22}{0.01} & \stdcell{0.88}{0.02} & \stdcell{0.47}{0.01} \\
\cmidrule(lr){2-9}
& \multirow{3}{*}{$50$}
& \textbf{H+E}   & \stdcell{0.55}{0.03} & \stdcell{0.49}{0.03} & \stdcell{0.28}{0.06} & \stdcell{0.18}{0.00} & \stdcell{\textbf{0.93}}{\textbf{0.00}} & \stdcell{\textbf{0.48}}{\textbf{0.00}} \\
&  & \textbf{H+O}   & \stdcell{\textbf{0.68}}{\textbf{0.04}} & \stdcell{\textbf{0.66}}{\textbf{0.04}} & \stdcell{\textbf{0.36}}{\textbf{0.09}} & \stdcell{\textbf{0.20}}{\textbf{0.00}} & \stdcell{0.87}{0.01} & \stdcell{0.47}{0.00} \\
&  & \textbf{H+E+O} & \stdcell{0.66}{0.06} & \stdcell{\textbf{0.66}}{\textbf{0.06}} & \stdcell{0.33}{0.12} & \stdcell{0.19}{0.00} & \stdcell{0.92}{0.02} & \stdcell{0.47}{0.00} \\
\cmidrule(lr){2-9}
& \multirow{3}{*}{$100$}
& \textbf{H+E}   & \stdcell{0.57}{0.04} & \stdcell{0.50}{0.04} & \stdcell{0.23}{0.06} & \stdcell{0.18}{0.00} & \stdcell{\textbf{0.93}}{\textbf{0.01}} & \stdcell{\textbf{0.48}}{\textbf{0.00}} \\
&  & \textbf{H+O}   & \stdcell{\textbf{0.66}}{\textbf{0.04}} & \stdcell{\textbf{0.66}}{\textbf{0.04}} & \stdcell{\textbf{0.33}}{\textbf{0.07}} & \stdcell{\textbf{0.20}}{\textbf{0.01}} & \stdcell{0.88}{0.02} & \stdcell{0.47}{0.00} \\
&  & \textbf{H+E+O} & \stdcell{\textbf{0.66}}{\textbf{0.04}} & \stdcell{\textbf{0.66}}{\textbf{0.04}} & \stdcell{0.32}{0.08} & \stdcell{0.19}{0.01} & \stdcell{\textbf{0.93}}{\textbf{0.02}} & \stdcell{0.47}{0.01} \\
\bottomrule
\end{tabular}}
\caption{\footnotesize
Feasibility results under controlled reveal levels.
$k_1$ and $k_2$ denote the percentages of experiments and outcomes revealed, and $R(\%) \in \{100, 50, 0\}$ denotes the reveal level. For each R, scenarios correspond to: $\textbf{H}: (k_1, k_2) = (0, 0)$; \textbf{H+E}: $(k1, k2) = (R, 0)$; \textbf{H+O}: $(k1, k2) = (0, R)$; \textbf{H+E+O}: $(k1, k2) = (R, R)$. For \textsc{Matter-of-Fact (MoF)} dataset, we report $Acc/F1_{macro}/MCC$ for label prediction and \textsc{Rouge} for rationale similarity (higher is better). Also note that each result averages 5 runs over the same 615 samples across all models (gpt-5.1, gpt-4o, Gemini-2.5-Pro/gem-pro, Gemini-2.5-Flash/gem-flash, Grok-4.1-fast/grok). See \autoref{fig:stability_curves} (in Appendix) for a compact visual summary of the main trends.}
\label{tab:main_std}
\end{table*}

\noindent \textsc{Matter-of-Fact} dataset \citep{jansen-etal-2025-matter} closely matches our formulation because each claim is paired
with a source paper whose experiments and outcomes may support, contradict, or
only partially address the hypothesis. Many claims depend on experimental scope,
intervention strength, or measurement conditions, reflecting realistic feasibility
scenarios in which a claim may be plausible but not empirically established. \textsc{Reasons} dataset \citep{saxena2025attributionscientificliteraturenew} provides a complementary positive-feasibility setting. Each instance
corresponds to a scientific statement explicitly supported by cited literature,
allowing us to examine whether models correctly recognize feasibility when valid
evidence exists. 

\noindent \textbf{Data leakage control.} To reduce pretraining leakage, we apply a model-specific post-cutoff filtering rule whenever a provider-documented knowledge cutoff is available. For \textsc{Matter-of-Fact}, this filtering is based on the benchmark's claim-level temporal metadata (\texttt{published\_date} / \texttt{exclude\_date}). In practice, the full \textsc{Matter-of-Fact} test split is naturally post-cutoff for GPT-4o (Oct.\ 2023), whereas later cutoffs such as GPT-5.1 (Sep.\ 2024) and Gemini-2.5-Pro/Flash (Jan.\ 2025) require narrower claim-specific subsets from the 2024--2025 test data. This protocol reduces, but does not eliminate, leakage risk, since papers may have circulated earlier as preprints, abstracts, or secondary summaries, and provider-side training details are not fully transparent.

\noindent \textbf{Controlled Evidence Construction.} For instances with an associated source
paper, we extract two structured evidence components: (i) experiment descriptions
$\mathcal{E}^\star$ and (ii) reported outcomes $\mathcal{O}^\star$. Extraction is performed once per
instance and reused across all models. We do not provide the full paper text to
avoid trivial claim matching and to isolate the effects of experimental design
and outcome information. The same extracted evidence is used across all settings (H, H+E, H+O, H+E+O).

\noindent\textbf{Extraction audit.} Because all evidence-conditioned evaluations depend on the extracted experiments and outcomes, we manually audited a targeted sample of extraction outputs against their source papers. The audit focused on two criteria: \textit{coverage}, i.e., whether the extracted items capture the key experiments and outcomes needed for feasibility judgment, and \textit{precision}, i.e., whether the extracted content is faithful to the source without adding unsupported details. Representative audit examples are provided in \autoref{tab:extraction_audit} (Appendix~\ref{app:extraction_audit}).

\noindent\textbf{Controlled prompting design.} Our goal is diagnostic evaluation rather than workflow engineering. We therefore use a minimal, fixed prompting setup so that performance differences can be attributed to the evidence provided rather than to added reasoning scaffolds or verification pipelines. Keeping the prompt structure constant across models, datasets, and evidence conditions helps isolate the effect of experiments and outcomes on feasibility judgments. 

\noindent \textbf{Partial-Information Stability Protocol:} To evaluate robustness under
incomplete evidence, we progressively remove portions of the extracted experiments and outcomes. For the previously defined reveal fractions $k_1$ and $k_2$ over $\{0, 0.5, 1.0\}$, we sample subsets uniformly without replacement for each configuration and evaluate them over five independent runs. Results are averaged to reduce sampling variance and to reveal non-monotonic degradation.

\section{Results and Analysis}
\label{sec:results}

Table~\ref{tab:main_std} summarizes feasibility performance under hypothesis-only
and controlled evidence settings and underpins the analyses that follow.

\noindent \textbf{RQ1: Feasibility from Internal Knowledge:} Under the hypothesis-only setting (\textsc{H}), all models achieve
non-trivial feasibility performance, indicating that parametric knowledge alone supports coarse feasibility judgments. However, performance variance across models is substantial, and justifications frequently rely on generic plausibility rather than explicit experimental constraints. This establishes \textsc{H} as a meaningful but brittle baseline: models can form priors about feasibility, but these priors are weakly grounded and sensitive to subsequent evidence.

\noindent \textbf{RQ2: Two Consistent Patterns in How Experiments and Outcomes Affect Feasibility}

\textit{Finding 1: Outcome evidence is more stabilizing than experiment descriptions.} Providing outcomes (\textsc{H+O}) improves or preserves feasibility
judgments more reliably than experiments alone (\textsc{H+E}). In contrast, \textsc{H+E} frequently degrades performance relative to \textsc{H}, particularly under partial evidence. This result is notable because experiments are often assumed to be the primary unit of scientific reasoning. Our findings suggest that, for current LLMs, experiment descriptions without results introduce ambiguity that models do not resolve reliably.

\textit{Finding 2: More evidence is not strictly better.}
In several settings, \textsc{H+E+O} does not outperform \textsc{H+O}, and in some cases underperforms it. This indicates that models do not consistently integrate experiments and outcomes compositionally; instead, additional context can interfere with decision-making.

\noindent \textbf{RQ3: Robustness Under Partial Evidence:} We next analyze robustness using progressive evidence removal. If models aggregated evidence coherently, performance would degrade smoothly as experiments or outcomes are removed.

To see this pattern clearly, we compare the hypothesis-only baseline (H) against the 50\% and 100\% reveal conditions for H+E, H+O, and H+E+O in \autoref{tab:main_std} (also shown in Appendix \autoref{fig:stability_curves}). The results show that performance does not drop smoothly as evidence is held back: in several cases, the 50\% condition performs worse than the 100\% version, and sometimes even worse than using the hypothesis alone. This directly shows that partial scientific context can mislead the model, not just give it less to work with.

\textit{Finding 3: Degradation is non-monotonic and sometimes adversarial:} Across models, we observe non-monotonic degradation: intermediate reveal levels (e.g., 50\%) often perform worse than both full evidence and no evidence. Moreover, a substantial fraction of partial-evidence configurations fall below the \textsc{H} baseline. This behavior demonstrates that partial scientific context can be \emph{actively misleading}, not merely insufficient. Such effects are invisible in standard full-context evaluations.

\noindent \textbf{Explaining the Failure Modes:} Our results point to three systematic limitations in current LLM-based feasibility assessment. (a) \textit{Surface alignment over evidential validity:} Models often overweight lexical or topical alignment between hypotheses and experiment descriptions, even when critical variables (e.g., intervention strength, endpoints, controls) do not match. Outcome statements mitigate this failure by imposing explicit directional
constraints that are harder to rationalize away. (b) \textit{Lack of evidence relevance gating:} When provided with mismatched or incomplete evidence, models rarely question
whether the evidence actually tests the hypothesis.
Instead, they attempt to reconcile any provided context, leading to brittle or incorrect feasibility judgments. (c) \textit{Anchoring under partial information:} Under partial evidence, models appear to anchor on the available fragment and
overcommit, rather than reverting to uncertainty. This explains why partial evidence can degrade performance more than no evidence at all.

\section{Conclusion}
\label{conclusion}
Our findings suggest that current LLMs treat feasibility as surface-level classification rather than as evidence-conditioned judgment. Improving feasibility assessment will require mechanisms that explicitly model evidence relevance, uncertainty, and variable alignment, rather than relying on additional context alone. Importantly, our controlled evidence framework reveals failure modes that are not detectable in standard claim-verification or retrieval-augmented settings, highlighting the need for diagnostic evaluations when deploying LLMs in scientific workflows.

\section*{Limitations}
\textbf{Protocol granularity.}
Our stability analysis uses three controlled reveal levels ($k_1,k_2 \in \{0,0.5,1.0\}$) with uniform subsampling to isolate when partial evidence helps versus misleads. Finer-grained reveal schedules and structured omission regimes (e.g., retaining only the most decision-critical outcomes) are natural extensions.

\textbf{Residual pretraining leakage risk.} Although we use post-cutoff paper filtering for models with documented knowledge cutoffs, leakage risk is not zero. Some source papers may have circulated earlier through preprints, abstracts, or other public summaries, and provider training corpora are not fully transparent. Accordingly, our evaluation should be interpreted as reducing leakage risk under a conservative post-cutoff protocol, rather than proving the absence of memorized exposure.

\textbf{Scope of models and benchmarks.}
We study a small set of widely used API-accessible LLMs on datasets that provide claim-level feasibility labels and associated source evidence. Our conclusions are therefore best interpreted as characterizing \emph{these} models and \emph{these} evidence formats under a controlled-evidence evaluation, rather than as an exhaustive survey of scientific domains or model families (e.g., open-weight systems).

\textbf{Task and metric choices.} We model feasibility as a binary decision to support controlled comparisons across evidence conditions while holding the prediction task fixed. This design is intentional: our goal is not to fully model the uncertainty, conditionality, or partial support that often characterize real scientific judgment, but to isolate how LLM predictions change as evidence availability and type vary. This formulation is also aligned with the annotation schema of the datasets used in our study, which provides claim-level \textsc{feasible}/\textsc{infeasible} labels. Introducing additional categories, such as partially supported, uncertain, or conditionally feasible, would add additional sources of variation that are difficult to disentangle from the main factor under study, namely the evidence provided to the model. In this setting, binary labels are especially useful because label changes provide an unambiguous signal of evidence sensitivity under controlled evidence conditions. At the same time, we acknowledge that binary feasibility does not capture the full granularity of scientific reasoning and therefore limits the generalizability of our conclusions to richer expert decision settings. Extending the framework to multi-level or probabilistic feasibility judgments would be an important direction for future work, but would require annotations that are not available in the datasets used here.

\textbf{Controlled prompting.}
We maintain consistent task instructions across all models and datasets, ensuring that any changes in predictions are attributable to the evidence provided, rather than to variations in prompts.

\section*{Ethical considerations}
This work analyzes how Large Language Models (LLMs) assess scientific feasibility under controlled evidence settings (hypothesis-only vs.\ adding extracted experiments and/or outcomes, including partial reveal). Our study uses publicly available datasets and papers and does not involve human subjects or personally identifiable information. We emphasize that feasibility predictions and rationales are not authoritative scientific judgments: models may be brittle under incomplete or mismatched evidence and may produce confident but ungrounded explanations. We also acknowledge that LLM outputs can reflect biases from pre-training and the covered scientific domains, and we do not make normative claims about model decisions. Our goal is diagnostic---to characterize when evidence helps or harms---and to support more reliable evaluation and interpretation for domain experts, rather than to enable automated acceptance/rejection of scientific claims.

\section*{Acknowledgments}
This material is based on research that is in part supported by the DARPA for the SciFy program under agreement number HR00112520301. The U.S. Government is authorized to reproduce and distribute reprints for Governmental purposes notwithstanding any copyright notation thereon. The views and conclusions contained herein are those of the authors and should not be interpreted as necessarily representing the official policies or endorsements, either express or implied, of DARPA or the U.S. Government.

\bibliography{latex/custom}

\clearpage
\appendix
\section{Appendix}
\label{sec:appendix}

\subsection{Implementation Details}
\label{app:implementation}

All models are accessed via their public APIs with a low decoding temperature ($0.1$) to reduce stochasticity. For each setting, we use a single prompt template per model family and apply it unchanged across datasets. We include abstracted templates for (i) extracting experiments and outcomes from the source paper (\autoref{fig:prompt_extract_tests}) and (ii) feasibility prediction in the H+E+O setting (\autoref{fig:prompt_heo}), with outputs constrained to structured JSON. For reproducibility, we use five fixed random seeds (42, 123, 456, 789, 101112) for sampling and report results aggregated across seeds.
\begin{figure}[!ht]
\centering

\begin{tcolorbox}[
  enhanced,
  width=\linewidth,
  colback=gray!2,
  colframe=black!30,
  boxrule=0.3pt,
  arc=1mm,
  left=1.2mm, right=1.2mm,
  top=1.1mm, bottom=1.1mm,
  title={\footnotesize\textbf{Abstracted Prompt for Extracting Tests (Experiments + Outcomes)}},
  fonttitle=\footnotesize,
]

\footnotesize

\textbf{Input:} \texttt{\{hypothesis\}}, \texttt{\{paper\_text\}}.\\[0.35em]

\textbf{Task:} Extract the most relevant \textbf{experiments} and \textbf{outcomes} needed to judge whether the hypothesis is feasible.\\[0.35em]

\textbf{Rules:}
\begin{itemize}[leftmargin=*, itemsep=0.15em, topsep=0.15em]
  \item \textbf{Experiments} ($\leq$10): what was tested and what was measured (1--2 sentences each).
  \item \textbf{Outcomes} ($\leq$5): key results (direction + important limitation) (1--2 sentences each).
  \item Use \textbf{only} evidence reported in the provided paper text.
\end{itemize}
\vspace{0.25em}

\textbf{Output:} Return \textbf{ONLY} valid JSON:
\begin{tcblisting}{
  listing only,
  colback=white,
  boxrule=0pt,
  left=0mm, right=0mm, top=0.1mm, bottom=0mm,
  listing options={
    basicstyle=\ttfamily\footnotesize,
    columns=fullflexible
  }
}
{
  "experiments": ["..."],
  "outcomes": ["..."]
}
\end{tcblisting}

\end{tcolorbox}

\vspace{-0.8em}
\caption{Abstracted prompt for extracting experiments and outcomes from the source paper.}
\vspace{-1.0em}
\label{fig:prompt_extract_tests}
\end{figure}

\begin{figure}[!ht]
\centering
\begin{tcolorbox}[
  enhanced,
  width=\linewidth,
  colback=gray!2,
  colframe=black!30,
  boxrule=0.3pt,
  arc=1mm,
  left=1.2mm, right=1.2mm,
  top=1.1mm, bottom=1.1mm,
  title={\footnotesize\textbf{Abstracted Prompt for Feasibility (Hypothesis + Experiments + Outcomes)}},
  fonttitle=\footnotesize,
]
\footnotesize

\textbf{Input:} \texttt{\{hypothesis\}}, \texttt{\{experiments\}}, \texttt{\{outcomes\}}.\\[0.35em]

\textbf{Task:} Decide whether the hypothesis is \textsc{feasible} or \textsc{infeasible} given the provided experiments and outcomes.\\[0.35em]

\textbf{Rules:}
\begin{itemize}[leftmargin=*, itemsep=0.15em, topsep=0.15em]
  \item Base the decision on the provided evidence (do not add new experiments or results).
  \item If evidence is insufficient or conflicting, choose the best-supported label and state uncertainty in the explanation.
  \item \textbf{Explanation} (1--3 sentences): cite specific experiments/outcomes.
\end{itemize}
\vspace{0.25em}

\textbf{Output:} Return \textbf{ONLY} valid JSON:
\begin{tcblisting}{
  listing only,
  colback=white,
  boxrule=0pt,
  left=0mm, right=0mm, top=0.1mm, bottom=0mm,
  listing options={basicstyle=\ttfamily\footnotesize, columns=fullflexible}
}
{
  "decision": "feasible" or "infeasible",
  "explanation": "1--3 sentences referencing 
                the experiments/outcomes"
}
\end{tcblisting}

\end{tcolorbox}

\vspace{-0.8em}
\caption{Abstracted prompt for feasibility prediction given a hypothesis, experiments, and outcomes.}
\vspace{-1.0em}
\label{fig:prompt_heo}
\end{figure}

\newpage
\subsection{Manual Audit of Extracted Experiments and Outcomes and Stability Curves}
\label{app:extraction_audit}

Table~\ref{tab:extraction_audit} shows representative examples from our manual audit of extracted experiments and outcomes against their source papers. For each case, we report an abstracted summary of the extracted experiments and outcomes, together with whether the extraction captured the key evidence needed for feasibility judgment (\textit{coverage}), whether it remained faithful to the source without unsupported additions (\textit{precision}), and the main audit note or failure mode.

\begin{table*}[!ht]
\centering
\scriptsize
\setlength{\tabcolsep}{3pt}
\renewcommand{\arraystretch}{1.15}
\begin{tabularx}{\textwidth}{p{0.45cm} p{0.9cm} X X X p{0.8cm} p{0.8cm} X}
\toprule
\textbf{Case} & \textbf{Dataset} & \textbf{Claim / Hypothesis} & \textbf{Abstracted extracted experiments} & \textbf{Abstracted extracted outcomes} & \textbf{Coverage} & \textbf{Precision} & \textbf{Main audit note / failure mode} \\
\midrule
1 & MoF & The charge transfer resistance in lithium-ion batteries follows a U-shaped pattern with respect to state-of-charge, with higher resistances at very low and very high SOC values and lower resistances in the middle SOC range. (Claim ID: 	2412.10896v3\_4\_T) &
\textbullet~Full-cell cycling and impedance measurements across SOC.\newline
\textbullet~Symmetric-cell analyses to isolate electrode-specific effects.\newline
\textbullet~Temperature- and rate-dependent tests to separate charge-transfer contributions. &
\textbullet~Charge-transfer resistance shows a clear mid-SOC minimum.\newline
\textbullet~Resistance rises at both low and high SOC.\newline
\textbullet~The trend persists across temperature and aging, with graphite dominating at low SOC and NMC at high SOC. &
Pass & Pass & The extraction captures the key electrochemical measurements and the central U-shaped finding, while preserving the mechanistic interpretation rather than reducing it to a generic resistance trend. \\
\midrule
2 & MoF & Higher growth temperatures for Ta films on sapphire lead to improved structural and DC electrical properties but paradoxically result in worse microwave performance due to interface-related loss mechanisms. (Claim ID: 2412.16730v1\_10\_T)&
\textbullet~Ta films were grown on sapphire at multiple substrate temperatures.\newline
\textbullet~Structural characterization used XRD, TEM, and AFM.\newline
\textbullet~Electrical and superconducting properties were measured with transport and resonator experiments. &
\textbullet~Higher growth temperature improves crystallinity, surface quality, and DC/superconducting transport.\newline
\textbullet~Microwave resonator internal quality factor decreases at higher growth temperature.\newline
\textbullet~The degradation is attributed to lossy Ta/sapphire interfacial states or layers. &
Pass & Pass & The extraction covers both the improvement in structural/DC properties and the degradation in microwave behavior, which is essential to the claim’s paradoxical conclusion. \\
\midrule
3 & MoF & Substituting Si with Ge in LaRu$_3$Si$_2$ increases the superconducting transition temperature from 6.6\,K to 7.1\,K at $x=0.07$ Ge content. (Claim ID: 2503.22477v2\_0\_T) &
\textbullet~A LaRu$_3$Si$_{2-x}$Ge$_x$ substitution series was synthesized.\newline
\textbullet~Samples were characterized with XRD, resistivity, magnetization, and specific heat.\newline
\textbullet~Normal-state transport was examined across composition. &
\textbullet~Ge substitution is structurally incorporated successfully.\newline
\textbullet~Superconducting $T_c$ does not increase with Ge substitution.\newline
\textbullet~Bulk superconducting signatures shift to lower temperature, consistent with disorder-induced suppression. &
Pass & Pass & This is a strong negative-feasibility case: the extraction directly addresses the stated composition and temperature claim and preserves the contradiction clearly. \\
\midrule
4 & REA & Hardware-efficient ansatzes that ignore chemical information are unlikely to support large molecules effectively. (Claim ID: 2105.07127v1\_25) &
\textbullet~The paper compares hardware-efficient, UCCSD, QCC, and other chemically informed ansatzes.\newline
\textbullet~Experiments evaluate molecular VQE accuracy, circuit depth, and optimization behavior.\newline
\textbullet~Scaling with molecule size is also examined. &
\textbullet~Chemically informed ansatzes achieve better energy accuracy at lower depth.\newline
\textbullet~These ansatzes scale more favorably to larger molecules.\newline
\textbullet~Hardware-efficient ansatzes often require impractical depth and optimization effort as molecule size grows. (Paper ID: arXiv:1803.11173) &
Pass & Partial & The extraction supports the core claim well, but it is somewhat broader than the exact statement span and includes additional conclusions about optimization and scaling beyond the most local claim wording. \\
\midrule
5 & REA & Gumbel-Top-$k$ uses a softmax-based continuous relaxation of argmax/top-$k$ and achieves differentiability via reparameterization. (Claim ID: 2312.14474v1\_22) &
\textbullet~The paper studies Gumbel-Top-$k$ as a differentiable subset-selection / top-$k$ operator.\newline
\textbullet~It compares the method with non-differentiable and REINFORCE-style baselines.\newline
\textbullet~Experiments examine training behavior and task performance on ranking/subset-selection tasks. &
\textbullet~The method enables end-to-end gradient-based training through a reparameterized softmax relaxation.\newline
\textbullet~It reduces gradient variance relative to score-function estimators.\newline
\textbullet~It performs competitively but introduces a temperature-dependent approximation bias. &
Pass & Pass & The extraction preserves both the main methodological benefit and the caveat about approximation bias, making it a clean example of faithful evidence abstraction. \\
\bottomrule
\end{tabularx}
\caption{Manual audit summary of extracted experiments and outcomes. Coverage indicates whether the extracted evidence includes the main experiments and outcomes needed for feasibility judgment. Precision indicates whether the extraction is faithful to the source paper without unsupported additions. Claim ID denotes the identifier of the benchmark claim/example, which is associated with a source paper. MoF: \textsc{Matter-of-Fact}, REA: \textsc{Reasons}.}
\label{tab:extraction_audit}
\end{table*}

\label{app:stability_curves}
\begin{figure*}[!ht]
\centering
\begin{tikzpicture}
\begin{groupplot}[
    group style={
        group size=2 by 2,
        horizontal sep=1.6cm,
        vertical sep=1.9cm
    },
    width=0.41\textwidth,
    height=0.23\textwidth,
    xmin=0, xmax=100,
    xtick={0,50,100},
    grid=both,
    tick label style={font=\small},
    label style={font=\small},
    title style={font=\small},
    every axis plot/.append style={
        thick,
        mark size=2.8pt,
        line width=1.1pt
    },
    grid style={line width=0.25pt},
    major grid style={line width=0.25pt},
]

\nextgroupplot[
    title={MoF Accuracy},
    ylabel={Accuracy},
    ymin=0.52, ymax=0.69,
]
\addplot+[color=blue, mark=o] coordinates {(0,0.648) (50,0.550) (100,0.584)};
\addplot+[color=red, mark=square] coordinates {(0,0.648) (50,0.664) (100,0.656)};
\addplot+[color=teal, mark=triangle] coordinates {(0,0.648) (50,0.648) (100,0.654)};

\nextgroupplot[
    title={MoF MCC},
    ylabel={MCC},
    ymin=0.22, ymax=0.40,
]
\addplot+[color=blue, mark=o] coordinates {(0,0.382) (50,0.256) (100,0.274)};
\addplot+[color=red, mark=square] coordinates {(0,0.382) (50,0.344) (100,0.323)};
\addplot+[color=teal, mark=triangle] coordinates {(0,0.382) (50,0.319) (100,0.320)};

\nextgroupplot[
    title={REASONS Accuracy},
    xlabel={Reveal level (\%)},
    ylabel={Accuracy},
    ymin=0.82, ymax=0.94,
    legend columns=3,
    legend style={
        at={(0.5,-0.48)},
        anchor=north,
        draw=none,
        font=\small
    },
]
\addplot+[color=blue, mark=o] coordinates {(0,0.830) (50,0.914) (100,0.914)};
\addplot+[color=red, mark=square] coordinates {(0,0.830) (50,0.892) (100,0.902)};
\addplot+[color=teal, mark=triangle] coordinates {(0,0.830) (50,0.918) (100,0.924)};
\legend{$H{+}E$, $H{+}O$, $H{+}E{+}O$}

\nextgroupplot[
    title={REASONS Macro-F1},
    xlabel={Reveal level (\%)},
    ylabel={Macro-F1},
    ymin=0.44, ymax=0.48,
]
\addplot+[color=blue, mark=o] coordinates {(0,0.452) (50,0.468) (100,0.474)};
\addplot+[color=red, mark=square] coordinates {(0,0.452) (50,0.466) (100,0.466)};
\addplot+[color=teal, mark=triangle] coordinates {(0,0.452) (50,0.470) (100,0.470)};

\end{groupplot}
\end{tikzpicture}
\caption{Reveal-level stability curves derived from Table~\ref{tab:main_std}, where each point is averaged over the five evaluated models for the corresponding metric and reveal setting. Each curve starts from the shared hypothesis-only baseline $H$ at 0\% reveal and shows performance under 50\% and 100\% reveal for $H{+}E$, $H{+}O$, and $H{+}E{+}O$. These curves show that robustness depends strongly on evidence type: on MoF, $H{+}E$ is the most brittle condition, especially under partial reveal, whereas outcome-based settings are more stable; on REASONS, evidence augmentation is generally beneficial relative to $H$, though the gains are not uniform across settings.}
\label{fig:stability_curves}
\end{figure*}
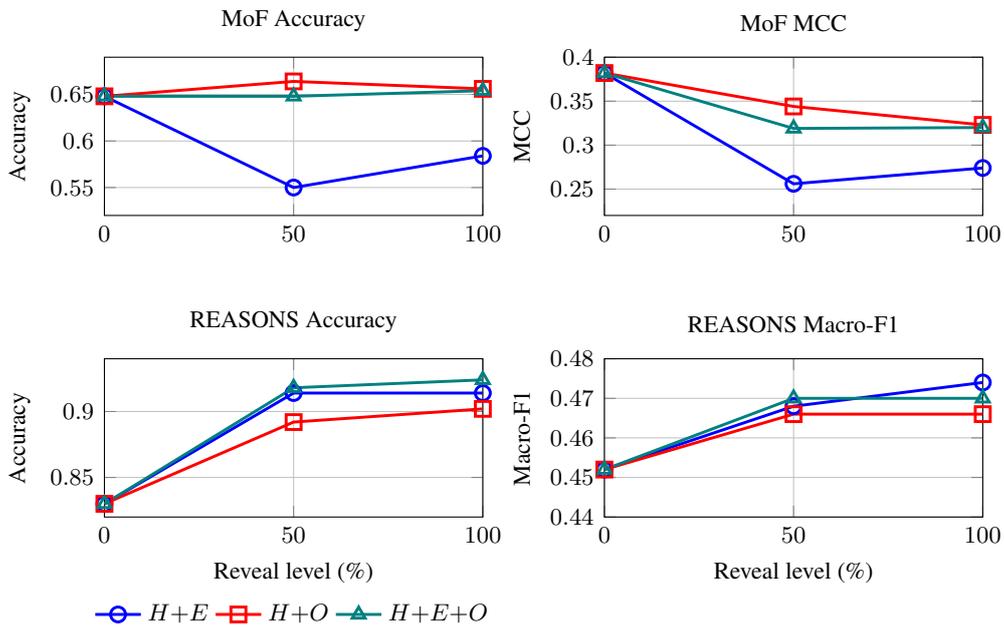

\end{document}